# Occupancy Grids: A Stochastic Spatial Representation for Active Robot Perception

## Alberto Elfes

Autonomous Robotics Laboratory
Department of Computer Sciences
IBM T. J. Watson Research Center
Yorktown Heights, NY 10598
Phone: (914)784-7944
E-mail (Internet): ELFES@IBM.COM

## Abstract

*In this paper we provide an overview of a new framework for robot perception, real-world modelling, and navigation that uses a stochastic tesselated representation of spatial information called the* Occupancy Grid. *The Occupancy Grid is a multi-dimensional random field model that maintains probabilistic estimates of the occupancy state of each cell in a spatial lattice. Bayesian estimation mechanisms employing stochastic sensor models allow incremental updating of the Occupancy Grid using multi-view, multi-sensor data, composition of multiple maps, decision-making, and incorporation of robot and sensor position uncertainty. We present the underlying stochastic formulation of the Occupancy Grid framework, and discuss its application to a variety of robotic tasks. These include range-based mapping, multi-sensor integration, path-planning and obstacle avoidance, handling of robot position uncertainty, incorporation of pre-compiled maps, recovery of geometric representations, and other related problems. The experimental results show that the Occupancy Grid approach generates dense world models, is robust under sensor uncertainty and errors, and allows explicit handling of uncertainty. It supports the development of robust and agile sensor interpretation methods, incremental discovery procedures, and composition of information from multiple sources. Furthermore, the results illustrate that robotic tasks can be addressed through operations performed directly on the Occupancy Grid, and that these operations have strong parallels to operations performed in the image processing domain.*

## 1 Introduction

Autonomous robot systems require the ability to recover robust spatial models of the surrounding world from sensory information and to efficiently utilize these models in robot planning and control tasks. These capabilities enable the robot to interact coherently with its environment, both by adequately interpreting the available sensor data so as to reach appropriate conclusions about the real world for short-term decisions, and by acquiring and manipulating a rich and substantially complete world model for long-term planning and decision-making.

Traditional approaches to robot perception have emphasized the use of geometric sensor models and heuristic assumptions to constrain the sensor interpretation process, and the use of geometric world models as the basis for planning robotic tasks [10, 5, 2]. "Low-level" sensing procedures extract geometric features such as line segments or surface patches from the sensor data, while "high-level" sensor interpretation modules use prior geometric models and heuristic assumptions about the environment to constrain the sensor interpretation process. The resulting deterministic geometric descriptions of the environment of the robot are subsequently used as the basis for other robotic activities, such as obstacle avoidance, path-planning and navigation, or planning of grasping and assembly operations. These approaches, which incorporate what we have characterized as the *Geometric Paradigm* in robot perception, have several shortcomings [5]. Generally speaking, the Geometric Paradigm leads to sparse and brittle world models; it requires early decisions in the interpretation of the sensor data for the instantiation of specific model primitives; it does not provide appropriate mechanisms for handling the uncertainty and errors intrinsic to the sensory information; and it relies heavily on the accurateness and adequacy of the prior world models and of the heuristic assumptions used. Overall, these approaches are of limited use in more complex scenarios, such as those encountered by mobile robots. Autonomous or semi-autonomous vehicles for planetary exploration, operation in hazardous environments, submarine exploration and servicing, mining and industrial applications have to explore and operate in unknown and unstructured environments, handle unforeseen events, and perform in real time.

In this paper, we review a new approach to robot perception and world modelling that uses a probabilistic tesselated representation of spatial information called the *Occupancy Grid* [8, 5, 10, 11]. The Occupancy Grid is a multi-dimensional random field that maintains stochastic estimates of the occupancy state of each cell in a spatial lattice. The cell estimates are obtained by interpreting sensor range data using probabilistic models that capture the uncertainty in the spatial information provided by the sensors. Bayesian estimation procedures allow the incremental updating of the Occupancy Grid using readings taken from several sensors and from multiple points of view. As a result, the disambiguation of sensor data is performed not through heuristics or prior models, but by additional sensing and through the use of adequate sensing strategies.

In subsequent sections, we provide an overview of the Occupancy Grid formulation and discuss how the Occupancy Grid framework provides a unified approach to a number of tasks in mobile robot perception and navigation. These tasks include range-based mapping, multiple sensor integration, path-planning and obstacle avoidance, handling of robot position uncertainty and other related problems. We show that a number of robotic problem-solving activities can be performed directly on the Occupancy Grid representation, precluding the need for the recovery of deterministic geometric descriptions. We also draw some parallels between operations on Occupancy Grids and related image processing operations.

## 2  The Occupancy Grid Framework

In this section, we provide a brief outline of the Occupancy Grid formulation, while in the succeeding sections we discuss several applications of the Occupancy Grid framework to mobile robot mapping and navigation. The scenarios under consideration in this paper involve a mobile robot operating in unknown and unstructured environments, and carrying a complement of sensors that provide range information directly (sonar, scanning laser rangefinders) or indirectly (stereo systems). A qualitative overview of some parts of this work is found in [11]; preliminary experimental results have been reported in [4, 8, 12], while a more detailed discussion is available in [10, 5]. More recently, we are applying the Occupancy Grid framework to the active control of robot perception [9] and as part of a multi-level performance-oriented mobile robot architecture [6].

### 2.1  The Occupancy Grid Representation

The Occupancy Grid is a multi-dimensional (typically 2D or 3D) tesselation of space into cells, where each cell stores a probabilistic estimate of its state. Formally, an *Occupancy Field* $O(\mathbf{x})$ can be defined as a discrete-state stochastic process defined over a set of continuous spatial coordinates $\mathbf{x} = (x_1, x_2, \ldots)$, while the *Occupancy Grid* is defined over a discrete spatial lattice. Consequently, the Occupancy Grid corresponds to a discrete-state (binary) random field [22]. A *realization* of the Occupancy Grid is obtained by estimating the state of each cell from sensor data.

More generally, the cell state can be used to encode a number of properties, represented using a random vector associated with each lattice point of the random field, and estimated accordingly. Properties of interest for robot planning could include occupancy, observability, reachability, connectedness, danger, reflectance, etc. We refer to such general world models, which are again instances of random fields, as *Inference Grids* [5]. In this paper, we are mainly interested in *spatial* models for robot perception, and will restrict ourselves to the estimation of a single property, the *occupancy state* of each cell.

### 2.2  Estimating the Occupancy Grid

In the Occupancy Grid, the state variable $s(C)$ associated with a cell $C$ is defined as a discrete random variable with two states, *occupied* and *empty*, denoted OCC and EMP. Since the states are exclusive and exhaustive, $P[s(C) = \text{OCC}] + P[s(C) = \text{EMP}] = 1$. Each cell has, therefore, an associated probability mass function that is estimated by the sensing process.

To construct a map of the robot's environment, two processing stages are involved. First, a sensor range measurement $r$ is interpreted using a stochastic sensor model. This model is defined by a probability density function (p.d.f.) of the form $p(r \mid z)$, where $z$ is the actual distance to the object being detected. Secondly, the sensor reading is used in the updating of the cell state estimates of the Occupancy Grid. For simplicity, we will derive the interpretation and updating steps for an Occupancy Grid defined over a single spatial coordinate, and outline the generalization to more dimensions.

In the continuous case, the random field $O(x)$ is described by a probability mass function defined for every $x$ and is written as $O(x) = P[s(x) = \text{OCC}](x)$, the probability of the state of $x$ being *occupied*. The probability of $x$ being *empty* is obviously given by $P[s(x) = \text{EMP}](x) = 1 - P[s(x) = \text{OCC}](x)$. The conditional probability of the state of $x$ being occupied given a sensor reading $r$ will be written as $O(x \mid r) = P[s(x) = \text{OCC} \mid r](x)$. For the discrete case, the Occupancy Grid corresponds to a sampling of the random field over a spatial lattice. We will represent the probability of a cell $C_i$ being occupied as $O(C_i) = P[s(C_i) = \text{OCC}](C_i)$, and the conditional probability given a sensor reading $r$ as $O(C_i \mid r) = P[s(C_i) = \text{OCC} \mid r](C_i)$. When only a single cell $C_i$ is being referenced, we will use the more succinct notation $P[s(C_i) = \text{OCC}]$.

We now consider a range sensor characterized by a sensor model defined by the p.d.f. $p(r \mid z)$, which relates the reading $r$ to the true parameter space range value $z$. Determining an optimal estimate $\hat{z}$ for the parameter $z$ is a straightforward estimation step, and can be done using Bayes' formula and MAP estimates [3, 21]. Recovering a model of the environment as a whole, however, leads to a more complex estimation problem. In general, obtaining an optimal estimate of the occupancy grid $O(C_i \mid r)$ would require determining the conditional probabilities of all possible world configurations. For the two-dimensional case of a map with $m \times m$ cells, a total of $2^{m^2}$ alternatives are possible, leading to a non-trivial estimation problem. To avoid this combinatorial explosion of grid configurations, the cell states are estimated as *independent* random variables. As a result, the Occupancy Grid corresponds to a Markov Random Field (MRF) of order 0 [22]. The independence assumption can be justified conceptually by the fact that in general there are no causal relationships between the occupancy states of different cells, and can be justified from an engineering point of view, because the resulting models are adequate for the range of tasks in which they are being applied. Finally, the computational simplicity intrinsic in the use of zero-order MRFs allows the development of very agile perception systems. On the other hand, there are applications, such as precise shape recovery, where more complex Occupancy Grid estimation models using higher-order MRFs are required [13, 15].

To determine how a sensor reading is used in estimating the state of the cells of the Occupancy Grid, we start by applying Bayes' theorem to a single cell $C_i$:

$$P[s(C_i) = \text{OCC} \mid r] = \frac{p[r \mid s(C_i) = \text{OCC}]\,P[s(C_i) = \text{OCC}]}{\sum_{s(C_i)} p[r \mid s(C_i)]\,P[s(C_i)]} \quad (1)$$

Notice that the $p[r \mid s(C_i)]$ terms that are required in this equation do not correspond directly to the sensor model $p(r \mid z)$, since the latter implicitly relates the range reading to the detection of a single object surface. In other words, the sensor model can be rewritten as:

$$p(r \mid z) = p[r \mid s(C_i) = \text{OCC} \wedge s(C_k) = \text{EMP}, k < i] \quad (2)$$

To derive the distributions for $p[r \mid s(C_i)]$, it is necessary to perform an estimation step over all possible world configurations. This can be done using Kolmogoroff's theorem [18]:

$$p[r \mid s(C_i) = \text{OCC}] = \sum_{\{G_{s(C_i)}\}} \big( p[r \mid s(C_i) = \text{OCC}, G_{s(C_i)}] \times$$
$$P[G_{s(C_i)} \mid s(C_i) = \text{OCC}] \big) \quad (3)$$

where $G_{s(C_i)} = (s(C_1) = s_1, \cdots, s(C_{i-1}) = s_{i-1}, s(C_{i+1}) = s_{i+1}, \cdots, s(C_n) = s_n)$ stands for a specific grid configuration with $s(C_i) = \text{OCC}$, and $\{G_{s(C_i)}\}$ represents all possible grid configurations under that constraint. In the same manner, $p[r \mid s(C_i) = \text{EMP}]$ can be computed as:

$$p[r \mid s(C_i) = \text{EMP}] = \sum_{\{G_{s(C_i)}\}} \big( p[r \mid s(C_i) = \text{EMP}, G_{s(C_i)}] \times$$
$$P[G_{s(C_i)} \mid s(C_i) = \text{EMP}] \big) \quad (4)$$

The configuration probabilities $P[G_{s(C_i)} \mid s(C_i)]$ are determined from the individual prior cell state probabilities. These, in turn, can be obtained from experimental measurements for the areas of interest, or derived from other considerations about likelihoods of cell states. We have opted for the use of non-informative or maximum entropy priors [1], which in this case assign equal probability values to the two possible states:

$$P[s(C_i) = \text{OCC}] = P[s(C_i) = \text{EMP}] = 1/2 \quad (5)$$

Using the cell independence assumption, these priors are used to determine the configuration probabilities $P[G_{s(C_i)} \mid s(C_i)]$, needed in Eqs. 3 and 4. Finally, Eq. 2 is used in the computation of the distributions $p[r \mid s(C_i)]$. The full derivation of these terms is found in [5]; we only remark that because there are subsets of configurations that are *indistinguishable* under a single sensor observation $r$, it is possible to derive closed form solutions of these equations for certain sensor models, and to compute numerical solutions in other cases.

To illustrate the approach, consider the case of an ideal sensor, characterized by the p.d.f. $p(r \mid z) = \delta(r - z)$, where $\delta$ is the Kronecker delta. For this case, the following closed form solution of Eq. 1 results (Fig. 1):

$$P[s(C_i) = \text{OCC} \mid r] = \begin{cases} 0 & \text{for } x < r, x \in C_i \\ 1 & \text{for } x, r \in C_i \\ 1/2 & \text{for } x > r, x \in C_i \end{cases} \quad (6)$$

which is an intuitively appealing result: if an ideal sensor measures a range value $r$, the corresponding cell has occupancy probability 1; the preceding cells are empty and have occupancy probability 0; and the subsequent cells have not been observed and are therefore unknown, having occupancy probability $1/2$.

As another example, consider a range sensor whose measurements are corrupted by Gaussian noise of zero mean and variance $\sigma^2$. The corresponding sensor p.d.f. is given by:

$$p(r \mid z) = \frac{1}{\sqrt{2\pi}\sigma} \exp\left(\frac{-(r-z)^2}{2\sigma^2}\right) \quad (7)$$

This equation can be used in the numerical evaluation of Eqs. 3 and 4. A plot of a typical cell occupancy profile obtained for this sensor from Eq. 1 is shown in Fig. 2.

To extend the derivation to two spatial dimensions, consider the example of a range sensor characterized by Gaussian uncertainty in both range and angle, given by the variances $\sigma_r^2$ and $\sigma_\theta^2$. In this case, the sensor p.d.f. can be represented in polar coordinates as:

$$p(r \mid z, \theta) = \frac{1}{2\pi\sigma_r\sigma_\theta} \exp\left[-\frac{1}{2}\left(\frac{(r-z)^2}{\sigma_r^2} + \frac{\theta^2}{\sigma_\theta^2}\right)\right] \quad (8)$$

In this formula, the dependency of the random variable $r$ on $z$ and $\theta$ is decoupled, a reasonable assumption for a first-order model of certain kinds of range sensors. Consequently, the estimation of the two-dimensional Occupancy Grid can be performed conveniently in polar coordinates $(\rho, \varphi)$, using fundamentally the same formulation as above (Eqs. 3 and 4) and applying Eq. 8 to recover the distributions $p[r \mid s(C_{\rho_i \varphi_i})]$. These in turn are used to obtain the polar Occupancy Grid $P[s(C_{\rho_i \varphi_i}) \mid r]$. To generate the corresponding two-dimensional cartesian Occupancy Grid, the polar grid can be scanned and resampled. A 2D cartesian Occupancy Grid is shown in Fig. 3, obtained from a single sonar reading. Similar derivations can be performed for 3D Occupancy Grids.

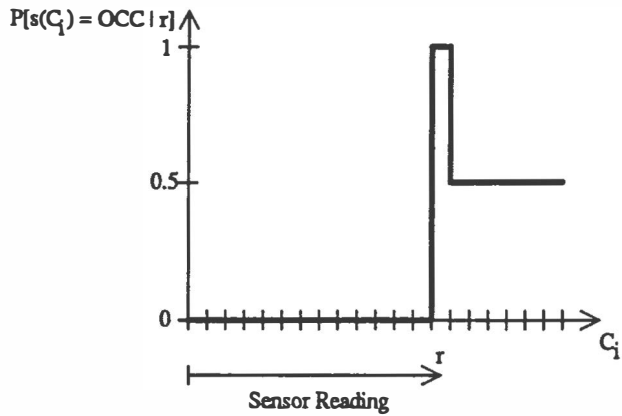

**Figure 1:** Occupancy Probability Profile for an ideal sensor, given a range measurement $r$.

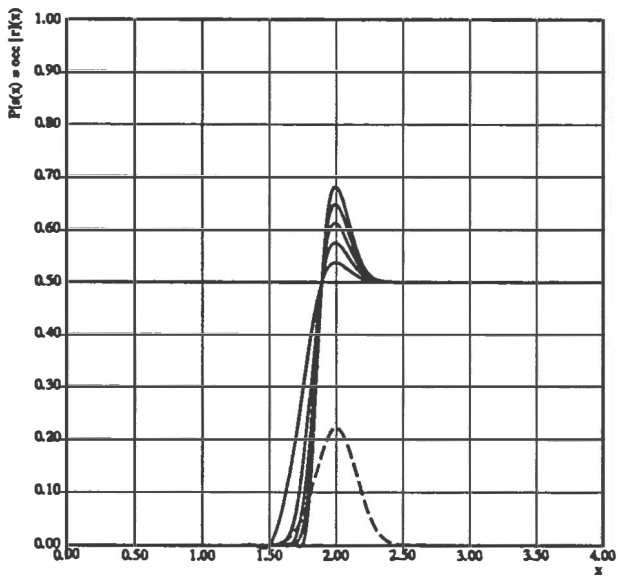

**Figure 2:** Occupancy Probability Profiles obtained from a sensor with Gaussian distribution. The sensor model $p(r \mid z)$ is shown superimposed (dashed line). Several successive updates of the cell occupancy probabilities are plotted, with the sensor positioned at $x = 0.0$ and with $r = 2.0$. The grid was initialized with $P[s(x) = \text{OCC}](x) = 0.5$. The profiles show that the Occupancy Grid converges towards the behaviour of the ideal sensor.

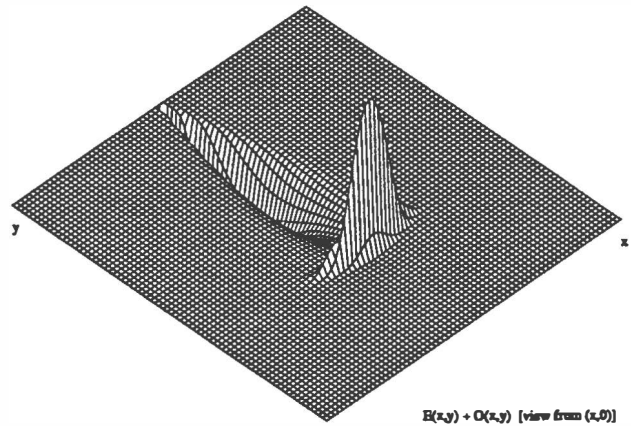

**Figure 3:** Two-Dimensional Sonar Occupancy Grid. The occupancy profile shown corresponds to a range measurement taken by a sonar sensor positioned at the upper left, pointing to the lower right. The plane corresponds to the UNKNOWN (1/2) level.

The sensor model actually used in most of our experiments, and that can be tailored to a large class of range sensors, is expressed as

$$p(\mathbf{r} \mid \mathbf{z}) = D(\mathbf{z})G(\mathbf{r}, \mathbf{z}, \Sigma(\mathbf{z})) \tag{9}$$

where, in addition to the multivariate Gaussian range measurement noise encoded in $G$, we also take into account the detection probability of an object at distance $\mathbf{z}$, $D(\mathbf{z})$, and the dependency of the range covariance on the object distance, expressed as $\Sigma(\mathbf{z})$. These terms were obtained through calibration experiments performed in our laboratory.

### 2.3 Updating the Occupancy Grid

Due to the intrinsic limitations of sensor systems, recovering a description of the world from sensory information is fundamentally an underconstrained problem. As mentioned previously, this has historically been addressed by the heavy use of prior models and simplifying heuristic assumptions about the robot's environment, leading to slow and brittle systems. Within the Occupancy Grid framework, the underconstrainedness of the sensor data is handled instead by the use of active perception strategies to resolve sensor ambiguity and uncertainty. Rather than relying on a single observation to obtain an estimate of the Occupancy Grid, information from multiple sensor readings taken from different viewpoints is composed to incrementally improve the sensor-derived map. This leads naturally to an emphasis on higher sensing rates and on the development of adequate sensing strategies.

To allow the incremental composition of sensory information, we use the sequential updating formulation of Bayes' theorem [5]. Given the current estimate of the state of a cell $s(C)$, $P[s(C_i) = \text{OCC} \mid \{r\}_t]$, based on observations $\{r\}_t = \{r_1, \cdots, r_t\}$, and given a new observation $r_{t+1}$, we can write:

$$P[s(C_i) = \text{OCC} \mid \{r\}_{t+1}] =$$
$$= \frac{p[r_{t+1} \mid s(C_i) = \text{OCC}] \, P[s(C_i) = \text{OCC} \mid \{r\}_t]}{\sum_{s(C_i)} p[r_{t+1} \mid s(C_i)] \, P[s(C_i) \mid \{r\}_t]} \quad (10)$$

In this formula, the previous estimate of the cell state, $P[s(C_i) = \text{OCC} \mid \{r\}_t]$, serves as the prior and is obtained directly from the Occupancy Grid. Tables for the sensor model-derived terms, $p[r_{t+1} \mid s(C_i)]$, can be computed offline for use in the recursive estimation procedure, allowing fast map updating. The new cell state estimate $P[s(C_i) = \text{OCC} \mid \{r\}_{t+1}]$ is subsequently stored again in the map. An example of this Bayesian updating procedure is shown in Fig. 2.

## 2.4 Sensor Integration

To increase the capabilities and the performance of robotic systems in general, a variety of sensing devices are necessary to support the different kinds of tasks to be performed. This is particularly important for mobile robots, where multiple sensor systems can provide higher levels of fault-tolerance and safety. Additionally, qualitatively different sensors have different operational characteristics and failure modes, and can therefore complement each other.

Within the Occupancy Grid framework, sensor integration can be performed using a formula similar to Eq. 10 for the combination of estimates provided by different sensors [5]. This allows the updating of the *same* Occupancy Grid by multiple sensors operating independently. Consider two independent sensors $S_1$ and $S_2$, characterized by sensor models $p_1(r \mid z)$ and $p_2(r \mid z)$. In this case, the integration of readings $r_{S_1}$ and $r_{S_2}$, measured by sensors $S_1$ and $S_2$, respectively, can be done using:

$$P[s(C_i) = \text{OCC} \mid r_{S_1}, r_{S_2}] =$$
$$= \frac{p[r_{S_2} \mid s(C_i) = \text{OCC}] \, P[s(C_i) = \text{OCC} \mid r_{S_1}]}{\sum_{s(C_i)} p[r_{S_2} \mid s(C_i)] \, P[s(C_i) \mid r_{S_1}]} \quad (11)$$

A different estimation problem occurs when separate Occupancy Grids are maintained for each sensor system, and integration of these sensor maps is performed at a later stage by composing the corresponding cell probability estimates. This requires the combination of probabilistic evidence from different sources [1]. Consider the two cell occupancy probabilities $P_1 = P_{S_1}[s(C_i) = \text{OCC} \mid \{r\}_{t_1}]$ and $P_2 = P_{S_2}[s(C_i) = \text{OCC} \mid \{r\}_{t_2}]$, obtained from separate Occupancy Grids built using sensors $S_1$ and $S_2$. The general solution to this problem involves the use of a *Superbayesian* approach [1]. For linear sensor performance evaluation models, the Superbayesian estimation procedure is reduced to a probabilistic evidence combination formula known as the *Independent Opinion Pool* [1]. Alternatively, the same result is obtained if a Bayesian integration is performed, with the use of maximum entropy priors. This method, when applied to the combination of the two sensor-derived estimates, $P_1$ and $P_2$, yields the simple formula [5]:

$$P[s(C_i) = \text{OCC} \mid P_1, P_2] = \frac{P_1 \, P_2}{P_1 \, P_2 + (1 - P_1) \, (1 - P_2)} \quad (12)$$

In previous work, described in [12, 16], the Independent Opinion Pool method was used to integrate Occupancy Grids derived separately from two sensor systems, a sonar array and a single-scanline stereo module, mounted on a mobile robot. An example of the resulting maps is presented in Section 3.2.

## 2.5 Incorporation of Pre-Compiled Maps

Throughout this paper we are mainly concerned with scenarios where the robot is operating in unknown environments, so that no pre-compiled maps can be used. There are other contexts, however, where such information *is* available. For example, mobile robots operating inside nuclear facilities could access detailed and substantially accurate maps derived from blueprints, while planetary rovers could take advantage of global terrain maps obtained from orbiting platforms. Such information can be represented using symbolic, topological and geometric models [14, 5]. The incorporation of these high-level pre-compiled maps can be done within the Occupancy Grid framework using the same methodology outlined in the previous sections. To provide a common representation, the geometric models are scan-converted into an Occupancy Grid, with occupied and empty areas being assigned the corresponding probabilities. These pre-compiled maps can subsequently be used as priors for sensor maps, or can simply be treated as another source of information to be integrated with sensor-derived maps [5].

## 2.6 Decision-Making

For certain applications, it may be necessary to make discrete choices concerning the state of a cell $C$. The *optimal estimate* is provided by the *maximum a posteriori* (MAP) decision rule [3], which can be written in terms of occupancy probabilities as:

$$\begin{cases} C \text{ is OCCUPIED} & \text{if } P(s(C) = \text{OCC}) > P(s(C) = \text{EMP}) \\ C \text{ is EMPTY} & \text{if } P(s(C) = \text{OCC}) < P(s(C) = \text{EMP}) \\ C \text{ is UNKNOWN} & \text{if } P(s(C) = \text{OCC}) = P(s(C) = \text{EMP}) \end{cases} \quad (13)$$

Additional factors, such as the cost involved in making different choices, can be taken into account by using other decision criteria, such as minimum-cost or minimum-risk estimates [21]. Depending on the specific application, it may also be of interest to define an UNKNOWN band, as opposed to a single thresholding value. As shown in [5], however, many robotic tasks can be performed directly on the Occupancy Grid, precluding the need to make discrete choices concerning the state of individual cells. In path-planning, for example, we define the cost of a path in terms of a risk factor directly related to the corresponding cell probabilities [8].

# 3  Using Occupancy Grids for Mobile Robot Mapping

We now proceed to illustrate the Occupancy Grid approach by discussing some applications of Occupancy Grids to autonomous mobile robots. In this section, we summarize the use of Occupancy Grids in sensor-based mobile robot *Mapping*, while in Section 4 we provide an overview of the use of Occupancy Grids in mobile robot *Navigation*. The experimental results shown here have been mostly obtained in operating environments that can be adequately described by two-dimensional maps. We have recently started to extend our work to the generation and manipulation of 3D Occupancy Grids [5].

One possible flow of processing for sensor-based robot mapping applications is outlined below and summarized in Fig. 4. As the mobile robot explores and maps its environment, the incoming sensor readings are interpreted using the corresponding probabilistic sensor models.The map of the world that the robot acquires from a single sensor reading is called a *Sensor View*. Various Sensor Views taken from a single robot position can be composed into *Local Sensor Maps*, which can be maintained separately for each sensor type. A composite description of the robot's surroundings is obtained through sensor integration of separate Local Sensor Maps into a *Robot View* (as mentioned previously, Robot Views can be generated directly from the integration of different sensors). As a result, the Robot View encapsulates the information recovered at a single mapping location. As the robot explores its surroundings, Robot Views taken from multiple data-gathering positions are composed into a *Global Map* of the environment. This requires relative registration of the Robot Views, an issue that is addressed in Section 4.

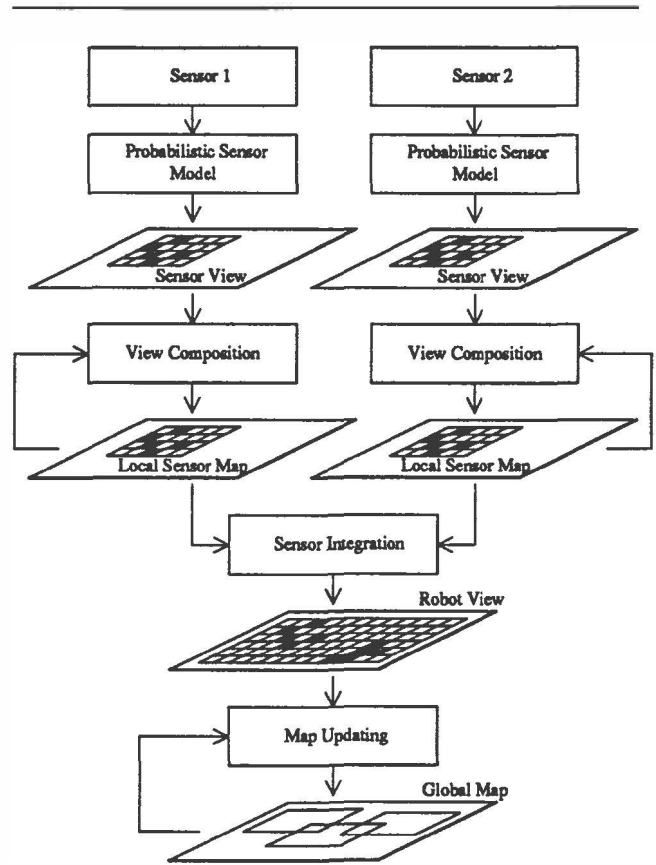

**Figure 4**: A Framework for Occupancy Grid Based Robot Mapping.

## 3.1  Sonar-Based Mapping

The Occupancy Grid representation was first developed in the context of sonar-based mapping experiments [4, 7, 8]. The functional limitations of sonar sensors and the need to recover robust and dense maps of the robot's environment precluded the use of simple geometric interpretation methods [8] and led to the investigation of tessellated probabilistic representations. Initial results using a heuristic approach called Certainty Grids [17, 4, 7, 8] were encouraging, and led to the development of the Occupancy Grid framework. To test the framework, we implemented an experimental system for sonar-based mapping and navigation for autonomous mobile robots called *Dolphin* [7, 8]. A number of indoor and outdoor experiments were performed (see, for example, [8, 5]). Fig. 5 presents a sonar map obtained during navigation down a corridor. The experimental work has shown that the cell updating mechanisms are computationally fast, allowing a high sensing to computation ratio, and that the framework can be equally well applied to other kinds of sensors [5, 12, 10].

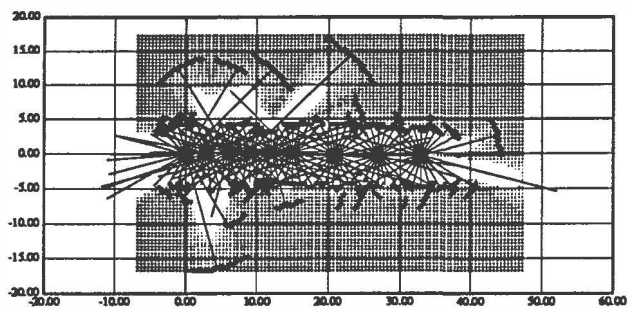

**Figure 5:** Sonar Mapping and Navigation Along a Corridor. Walls and open doors can be distinguished and the resolution is sufficient to allow even wall niches to be observed in the map. The range readings taken from each robot stop are drawn superimposed on the map.

### 3.2 Sensor Integration of Sonar and Scanline Stereo

The Occupancy Grid framework provides a straightforward approach to sensor integration. Range measurements from each sensor are converted directly to the Occupancy Grid representation, where data taken from multiple views and from different sensors can be combined naturally. Sensors are treated modularly, and separate sensor maps can be maintained concomitantly with integrated maps, allowing independent or joint sensor operation. In joint work with Larry Matthies, we have performed experiments in the integration of data from two sensor systems: a *sonar sensor array* and a *single-scanline stereo module* that provides horizontal depth profiles, both mounted on a mobile robot. This allows the generation of improved maps that take advantage of the complementarity of the sensors [12, 16]. A typical set of maps is shown in Fig. 6.

## 4 Using Occupancy Grids for Robot Navigation

For autonomous robot navigation, a number of concerns have to be addressed. In this section, we briefly outline the use of Occupancy Grids in path-planning and obstacle avoidance, estimating and updating the robot position, and incorporating the positional uncertainty of the robot into the mapping process (Fig. 7). A detailed discussion is found in [5].

### 4.1 Path-Planning and Obstacle Avoidance

In the *Dolphin* system, path-planning and obstacle avoidance are performed using potential functions and an A* search algorithm that operates directly on the Occupancy Grid. The path-planning operation minimizes a multi-objective cost function $f(\mathbf{P})$, defined over the path $\mathbf{P} = \{(x_0, y_0, \theta_0), \cdots, (x_n, y_n, \theta_n)\}$, that takes into account both the occupancy probabilities of the cells being traversed and the total distance to the robot's destination:

$$f(\mathbf{P}) = \tau_c \sum_{\forall C \in \mathbf{P}} \Gamma(C) + \tau_d \, \text{length}(\mathbf{P}) \qquad (14)$$

where $\tau_c$ and $\tau_d$ weigh the component costs that are associated with the cell occupancy probabilities and with the distance to the goal, respectively. The function $\Gamma(C)$ expresses the cost of traversing a single cell, and is de-fined directly as a non-linear function of the occupancy probability $P[s(C) = \text{OCC}]$ (see [5]).

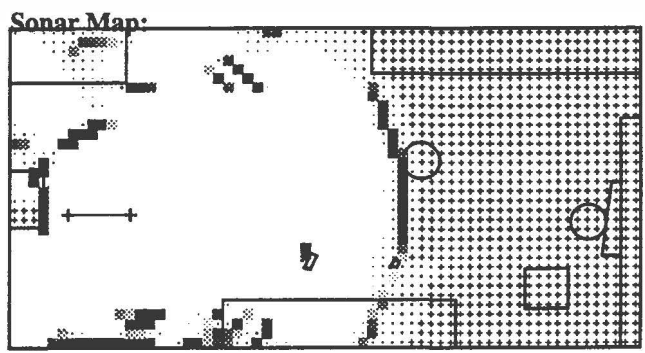

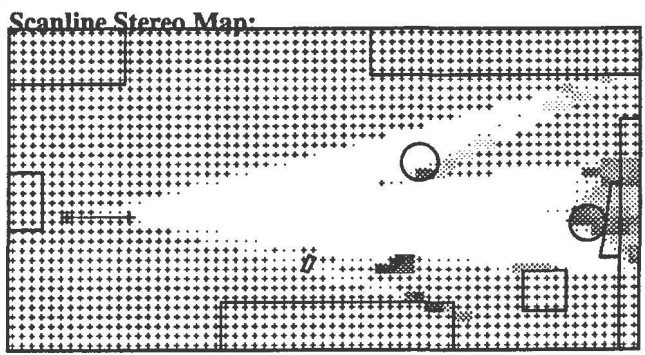

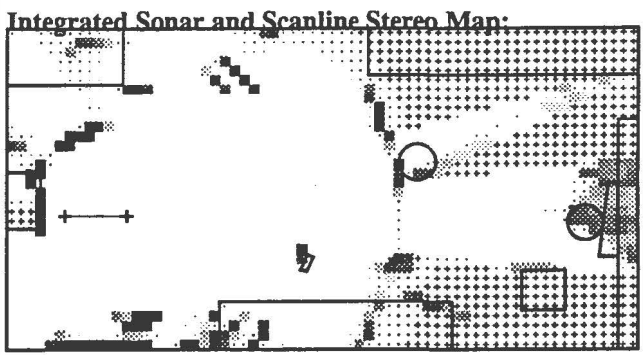

**Figure 6:** Sensor Integration of Sonar and Scanline Stereo. Occupancy Grids generated separately for sonar and scanline stereo, and jointly through sensor integration are shown. *Occupied* regions are marked by shaded squares, *empty* areas by dots fading to white space, and *unknown* spaces by + signs.

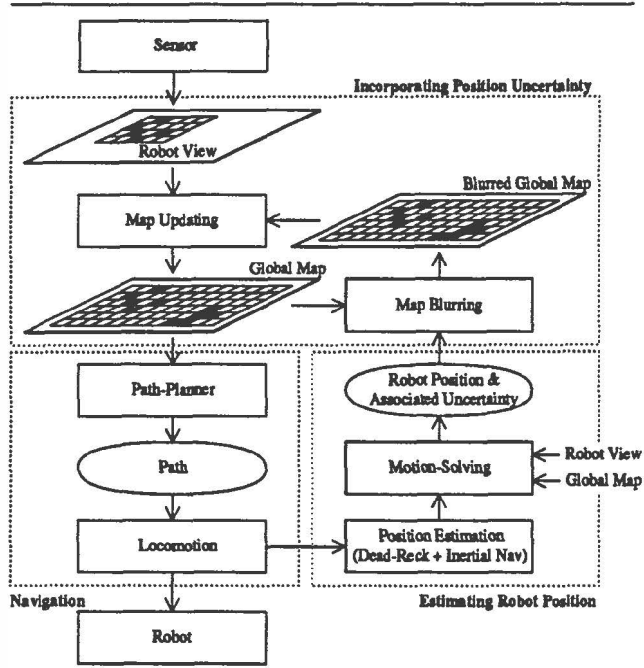

**Figure 7:** A Framework for Occupancy Grid-Based Robot Navigation.

### 4.2 Handling Robot Position Uncertainty

To desambiguate sensor information and recover accurate and complete descriptions of the environment of operation of a robot, it is necessary to integrate sensor data acquired from multiple viewing positions. To allow the composition of these multiple views into a coherent model of the world, accurate information concerning the relative transformations between data-gathering positions is necessary to allow precise registration of the views for subsequent integration. For mobile robots that move around in unstructured environments, recovering precise position information poses a major problem. Over longer distances, dead-reckoning estimates are not sufficiently reliable; consequently, motion-solving methods that use landmark tracking or map matching approaches are usually applied to reduce the registration imprecision due to motion. Furthermore, the positional error is compounded over sequences of movements as the robot traverses its environment. This leads to the need for explicitly handling positional uncertainty and taking it into account when composing sensor information.

To represent and estimate the robot position as the vehicle explores its environment, we use the *Approximate Transformation* (AT) framework [19]. A robot motion $M$, defined with respect to some coordinate frame, is represented as $\widetilde{M} = <\widehat{M}, \Sigma_M>$, where $\widehat{M}$ is the estimated (nominal) position, and $\Sigma_M$ is the associated covariance matrix that captures the positional uncertainty. The parameters of the robot motion are determined from dead-reckoning and inertial navigation estimates, which can be composed using the AT *merging* operation, while the updating of the robot position uncertainty over several moves is done using the AT *composition* operation [19].

### 4.3 Motion-Solving

For more precise position estimation, a multi-resolution correlation-based motion-solving procedure is employed. It searches for an optimal registration between the new Robot View and the current Global Map, by matching the corresponding Occupancy Grids before map composition [17].

### 4.4 Incorporating Positional Uncertainty into the Mapping Process

After estimating the registration between the new Robot View and the Global Map, the associated uncertainty is incorporated into the map updating process as a blurring or convolution operation performed on the Occupancy Grid. We distinguish between *World-Based Mapping* and *Robot-Based Mapping* [5, 11].

In *World-Based Mapping*, the motion of the robot is related to the observer or world coordinate frame, and the current Robot View is blurred by the robot's positional uncertainty prior to composition with the Global Map. If we represent the Global Map by $M_G$, the current Robot View by $V_R$, the robot position by the AT $\widetilde{R} = <\widehat{R}, \Sigma_R>$, the blurring operation by the symbol $\widetilde{\otimes}$ and the composition of maps by the symbol $\widetilde{\oplus}$, we can express the world-based mapping procedure as:

$$M_G \leftarrow M_G \widetilde{\oplus} (V_R \widetilde{\otimes} \widetilde{R}) \tag{15}$$

Since the global robot position uncertainty increases with every move, the effect of this updating procedure is that the new Views become progressively more blurred, adding less and less useful information to the Global Map. Observations seen at the beginning of the exploration are "sharp", while recent observations are "fuzzy". From the point of view of the inertial observer, the robot eventually "dissolves" in a cloud of probabilistic smoke.

For *Robot-Based Mapping* (Fig. 7), the registration uncertainty of the Global Map due to the recent movement of the robot is estimated, and the Global Map is blurred by this uncertainty prior to composition with the current Robot View. This mapping procedure can be expressed as:

$$M_G \leftarrow V_R \widetilde{\oplus} (M_G \widetilde{\otimes} \widetilde{R}) \tag{16}$$

A consequence of this method is that observations performed in the remote past become increasingly uncertain, while recent observations have suffered little blurring.

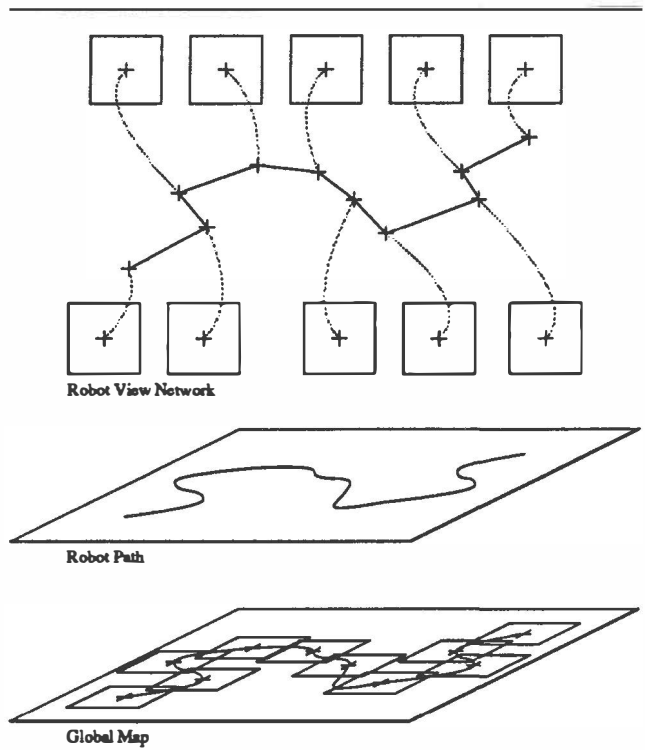

**Robot View Network**

**Robot Path**

**Global Map**

**Figure 9**: Maintaining a Dual Representation. A stochastic graph with the individual Robot Views is maintained in conjunction with the Global Map.

From the point of view of the robot, the immediate surroundings (which are of relevance to its current navigational tasks) are "sharp". The robot is leaving, so to speak, an expanding "probabilistic trail" of weakening observations behind it (see Fig. 8).

It should be noted, however, that the local spatial relationships observed within a Robot View still hold. So as not to lose this information, we use a two-level spatial representation, incorporating Occupancy Grids and Approximate Transformations. On one level, the individual Views are stored attached to the nodes of an AT graph (a *stochastic map* [20]) that describes the movements of the robot. Coupled to this, a Global Map is maintained that represents the robot's current overall knowledge of the world (Fig. 9).

## 5 Other Applications

In the previous sections, we have seen that Occupancy Grids provide a unified approach to a number of issues in Robotics and Computer Vision. Additional tasks that we have addressed include the recovery of geometric descriptions from Occupancy Grids [7, 8], incorporation of pre-compiled maps [5], landmark recognition, prediction of sensor readings from Occupancy Grids, and related prob-lems. We are currently extending this work in several directions; these include the generation of 3D Occupancy Grids from depth profiles derived from laser scanners or stereo systems, detection of moving objects using space-time filtering techniques, development of a methodology for active control of robot perception [9], and the incorporation of the Occupancy Grid framework in a multi-level performance-oriented mobile robot architecture [6].

| Comparison of Operations on Occupancy Grids and on Images | |
|---|---|
| Occupancy Grids | Images |
| Labelling cells as Occupied, Empty or Unknown | Thresholding |
| Handling Position Uncertainty | Blurring/Convolution |
| Removing Spurious Spatial Readings | Low-Pass Filtering |
| Motion Solving/Map Matching | Correlation |
| Obstacle Growing for Path-Planning | Region Growing |
| Path-Planning | Edge Tracking |
| Determining Object Boundaries | Edge Detection |
| Extracting and Labelling Occupied and Empty Areas | Segmentation/Region Colouring/Labelling |
| Prediction of Sensor Readings from User-Provided Maps | Convolution |
| Incorporating User-Provided Maps | Sum-Conversion |
| Object Motion Detection over Map Sequences | Space-Time Filtering |

**Figure 10**: An Overview of Operations on Occupancy Grids and the Corresponding Image Processing Operations.

It should be noted that many robotic tasks can be performed on Occupancy Grids using operations that are similar or equivalent to computations performed in the image processing domain. Table 10 provides a qualitative overview and comparison of some of these operations.

We finalize our remarks with a note concerning low-level versus high-level representations. It is interesting to observe that in Robotics and Computer Vision there has been historically a slow move from very high-level (stylized) representations of blocks-world objects to the recovery of simple spatial features in very constrained real images; from there to the recovery of surface patches; and recently towards "dense", tesselated representations of spatial information such as the Occupancy Grid. A parallel evolution from sparse, high-level or exact descriptions to dense, lower-level and sometimes approximate descriptions can be seen in some other computational fields, such as Computer Graphics and Finite Element Analysis.

## 6 Conclusions

We have reviewed in this paper the Occupancy Grid framework and presented results from its application to mobile robot mapping and navigation tasks in unknown and unstructured environments. The Occupancy Grid approach supports agile and robust sensor interpretation methods, incremental discovery procedures, composition of information from multiple sensors and over multiple positions of the robot, and explicit handling of uncertainty. Furthermore, the world models recovered from sensor data can be used efficiently in robotic planning and problem-solving activities. The results lead us to suggest that the Occupancy Grid framework provides a novel approach to robot perception and spatial reasoning that has the characteristics of robustness and generality necessary for real-world robotic applications.

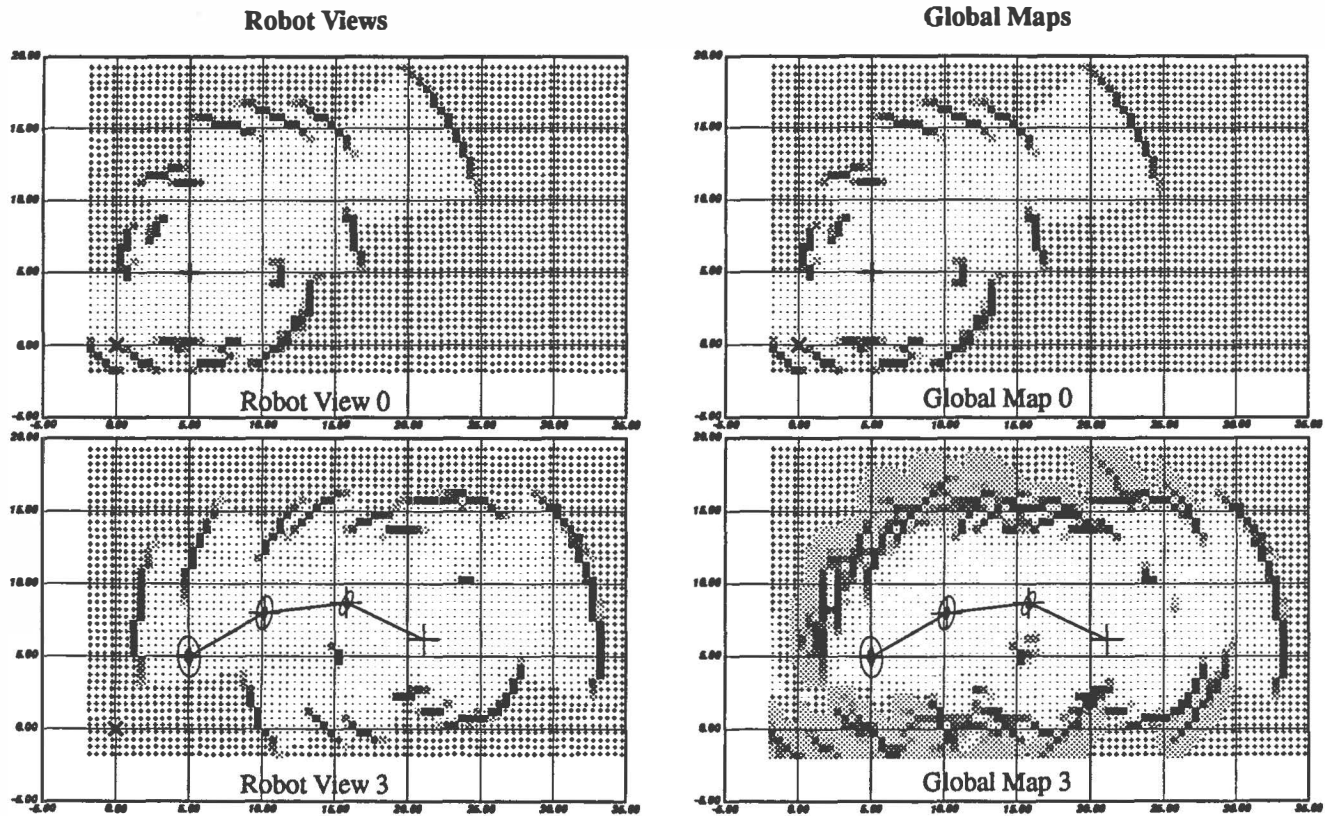

**Figure 8:** Incorporating Motion Uncertainty into the Mapping Process. For robot-centered mapping, the Global Map is blurred by the robot position uncertainty (shown using the corresponding covariance ellipses) prior to composition with the Robot View. Two stages of the process are shown.

## Acknowledgments

The author wishes to thank Hans Moravec, Sarosh Talukdar, Peter Cheeseman, Radu Jasinschi, Larry Matthies, José Moura, Michael Meyer and Larry Wasserman for their comments and suggestions concerning some of the issues discussed in this paper.

Most of the research discussed in this paper was performed when the author was with the Mobile Robot Lab, Robotics Institute, Carnegie-Mellon University, and with the Engineering Design Research Center, CMU. It was supported in part by the Office of Naval Research under Contract N00014-81-K-0503. The author was supported in part by a graduate fellowship from the Conselho Nacional de Desenvolvimento Científico e Tecnológico - CNPq, Brazil, under Grant 200.986-80, in part by the Mobile Robot Lab, CMU, and in part by the Engineering Design Research Center, CMU.

The views and conclusions contained in this document are those of the author and should not be interpreted as representing the official policies, either expressed or implied, of the funding agencies.